\documentclass[10pt,journal,compsoc]{IEEEtran}
\usepackage[utf8]{inputenc}
\usepackage{adjustbox}
\usepackage{tabularx}
\usepackage{graphicx,color}
\usepackage{balance}
\usepackage[labelformat=simple, labelsep=colon]{subcaption}
\usepackage[dvipsnames]{xcolor} %

\usepackage[compact]{titlesec}         
\titlespacing{\section}{2pt}{2pt}{2pt} 
\AtBeginDocument{
  \setlength\abovedisplayskip{2pt}
  \setlength\belowdisplayskip{2pt}}

\usepackage[rawfloats=true]{floatrow} 
\restylefloat{figure}     

\title{Edge Intelligence for Empowering IoT-based Healthcare Systems}

\author{
\IEEEauthorblockN{\textbf{Vahideh Hayyolalam}, \textit{Member, IEEE}, \textbf{Moayad Aloqaily},
\textit{Member, IEEE}, \textbf{\"Oznur \"Ozkasap}, \textit{Senior Member, IEEE}, \textbf{Mohsen Guizani}, \textit{Fellow, IEEE }
}\vspace{-4pt}
\IEEEcompsocitemizethanks{

\IEEEcompsocthanksitem V. Hayyolalam and \"O. \"Ozkasap are with Ko\c{c} University, Istanbul, Turkey. \protect E-mails: \{vhayyolalam20; oozkasap\}@ku.edu.tr.

\IEEEcompsocthanksitem M. Aloqaily is with xAnalytics Inc., Canada. \protect E-mail: maloqaily@xanalytics.ca.

\IEEEcompsocthanksitem M. Guizani is with Qatar University, Qatar. \protect E-mail: mguizani@ieee.org.

}
}

\begin{document}
\maketitle
\begin{abstract}
The demand for real-time, affordable, and efficient smart healthcare services is increasing exponentially due to the technological revolution and burst of population. To meet the increasing demands on this critical infrastructure, there is a need for intelligent methods to cope with the existing obstacles in this area. In this regard, edge computing technology can reduce latency and energy consumption by moving processes closer to the data sources in comparison to the traditional centralized cloud and IoT-based healthcare systems. In addition, by bringing automated insights into the smart healthcare systems, artificial intelligence (AI) provides the possibility of detecting and predicting high-risk diseases in advance, decreasing medical costs for patients, and offering efficient treatments. The objective of this article is to highlight the benefits of the adoption of edge intelligent technology, along with AI in smart healthcare systems. Moreover, a novel smart healthcare model is proposed to boost the utilization of AI and edge technology in smart healthcare systems. Additionally, the paper discusses issues and research directions arising when integrating these different technologies together.
\end{abstract}


\section{Introduction}
    Healthcare is one of the crucial parts of human life. Due to the rapid population growth and the rise of various diseases, there exists a drastic surge in requests for healthcare and its facilities. Also, with the technological evolution and the emergence of Internet of things (IoT) coupled with the improvement of next-generation wireless communications, the concept of smart healthcare or connected healthcare systems has been introduced as a developed version of conventional healthcare systems. In fact, smart healthcare refers to a health service that exploits technologies such as IoT, wearable devices, and advanced communication protocols to dynamically connect patients, caregivers, and health institutions and transfer information among them \cite{abdellatif2019edge}. Smart healthcare services provide the possibility of handling and responding to medical requests remotely in an intelligent way, which diminishes hospitalization remarkably and helps people and doctors to predict, detect, diagnose, and treat diseases intelligently. Also, smart healthcare is used for preventing and controlling the outbreak of contagious and infectious diseases such as the Ebola virus, Avian influenza, Chickungunya virus \cite{rani2018smart}, and recently the Covid-19 pandemic. Taking all the above mentioned issues into account, adoption of efficient smart healthcare services would certainly improve the health level of our society.
    
    In order to expand the  healthcare facilities and satisfy the vast amount of users’ requests, smart healthcare systems employ numerous smart devices and IoT technologies. Thus, a massive amount of heterogeneous medical records generated by smart devices need to be processed and evaluated accurately based on the target of the system \cite{tuli2020healthfog}. Therefore, there is a need for effective communications among various entities of smart healthcare community and data centers to ensure low response time for emergency cases in real-time health needs, since the high response time and high latency at data centers are critical risks in smart healthcare systems and can lead to the irreparable catastrophes \cite{abdellatif2019edge}. Keeping these facts in mind, edge computing technology, and AI are the best choices for these issues. Especially, adopting the combination of these two technologies can pave the way for solving various challenges in the scope of smart healthcare.
    
    Edge computing technology moves some processes close to the data sources, which reduces the load of transmitted data conspicuously \cite{pathinarupothi2018iot}. In fact, edge technology enables smart healthcare systems to conduct some processes and store some data near the end-users rather than transferring all the records to the cloud data centers \cite{azar2019energy}. In this way, smart healthcare systems only need to send the results of processes and some raw data to the remote cloud data centers. Various research works by utilizing edge technology in the field of smart healthcare can conduct big data analysis, 
    store and process sensitive medical data, 
    diminish latency, 
    reduce energy consumption, 
    reduce cost, 
    reduce network congestion, and response time. 

    Moreover, by emulating human cognition in analyzing data through complex algorithms, AI can smooth the path of estimating the conclusions without the direct involvement of human. In fact, AI is responsible for investigating relations between the treatment, prevention, or detection techniques of diseases based on the collected data from patients 
    In the edge context, the resource constraints necessitate innovative and lightweight methods to be able to execute them at the edge environment. Therefore, Machine Learning (ML) and Deep Learning (DL) as the subsets of AI are mostly preferred in edge technology to train the system for learning, making decisions, and actuating. The integration of edge computing and AI introduces an edge intelligence concept, which leads to emerge a revolution in smart healthcare systems \cite{al2019lightweight}. Researchers have proposed several smart healthcare systems adopting AI methods such as DL, 
    Deep Neural Network (DNN), 
    reinforcement learning (RL), 
    K-means 
    , and other ML-based algorithms. 
   
   
 The contributions of this article can be summarized as follows: \begin{itemize}
        \item Classifies  the  current  AI-assisted  edge-based  smart healthcare  solutions  via  three viewpoints,  namely, the  application  cases  in  IoT  smart  healthcare,  the proper AI deployment location at the edge, and the most suitable type of adopted AI technique.
        \item Proposes a novel smart healthcare model leveraging AI and edge technology together to reduce the limitations of the existing models.
        \item Adopts the use of deep reinforcement learning (DRL). According  to  the number  of  vital sign types needed to be monitored and the amount of collected medical data, the  proposed  system  is designed to distribute the DRL layers and processes optimally.
        \item Generalize the design of the proposed DRL model in a way it can be used in parallel mode at the edge layer. It means that the processes with the same functionality are distributed among different devices or DRL layers to reduce response time, latency, and congestion.
    \end{itemize}
    
    
    The magazine is organized as follows, Section \ref{Sec.emergingEdge} discusses how edge computing affects smart healthcare systems. Then, Section \ref{Sec.AI} states the AI implication in edge-based smart healthcare systems as well as the pros and challenges of integrating AI and edge technology. Section \ref{sec.frame} introduces a new framework for smart healthcare exploiting AI and edge technology. Open issues and challenges are discussed in Section \ref{sec.openissues}. Eventually, Section \ref{sec.conclusion} provides the conclusion of this research. 
    
\section{Emerging edge technology in smart healthcare} \label{Sec.emergingEdge}
    Many healthcare systems have adopted cloud computing to deliver affordable and scalable solutions in order to process and store massive amounts of data recorded via numerous biosensors. However, since cloud-oriented healthcare systems mostly include mobile devices, network, and cloud servers, there are often long distances between the systems’ units, which intensifies a high latency issue in these systems \cite{azar2019energy}. Thus, emergency and real-time medical demands cannot rely on cloud-based healthcare systems. Furthermore, the vast amounts of records produced by sensors need to regularly be transferred to the cloud for being processed and stored, which leads to high energy consumption and high cost. Also, most of chronic patients require low-cost mobile environments, which is not supported by cloud-oriented solutions \cite{tuli2020healthfog}.

   Edge computing as an additional layer between end-users’ devices and remote cloud data centers is an emerging solution to prevail cloud-based healthcare limitations. In fact, edge computing is a smart gateway that plays the role of a bridge between devices and cloud data servers \cite{hosseini2020multimodal}. Edge-assisted solutions transfer processes close to the end-users, which provides short response time and reduces energy usage. Indeed, edge technology enables smart healthcare systems to mine and process the collected data on the edge servers and devices near the user rather than transferring it to the cloud servers. Processing at the edge also assists and boosts security, privacy, mobility, low network bandwidth usage, geographical distribution, and location awareness and facilitates online diagnosis and analytics, which reduces clinic and hospital visits \cite{tuli2020healthfog}. Hence, edge technology contributes to the development of smart healthcare services by providing swift, more comprehensive, and ubiquitous treatments.
     \begin{figure} [htbp!] 
    \centering
    \includegraphics[height=110 mm, width=90mm]{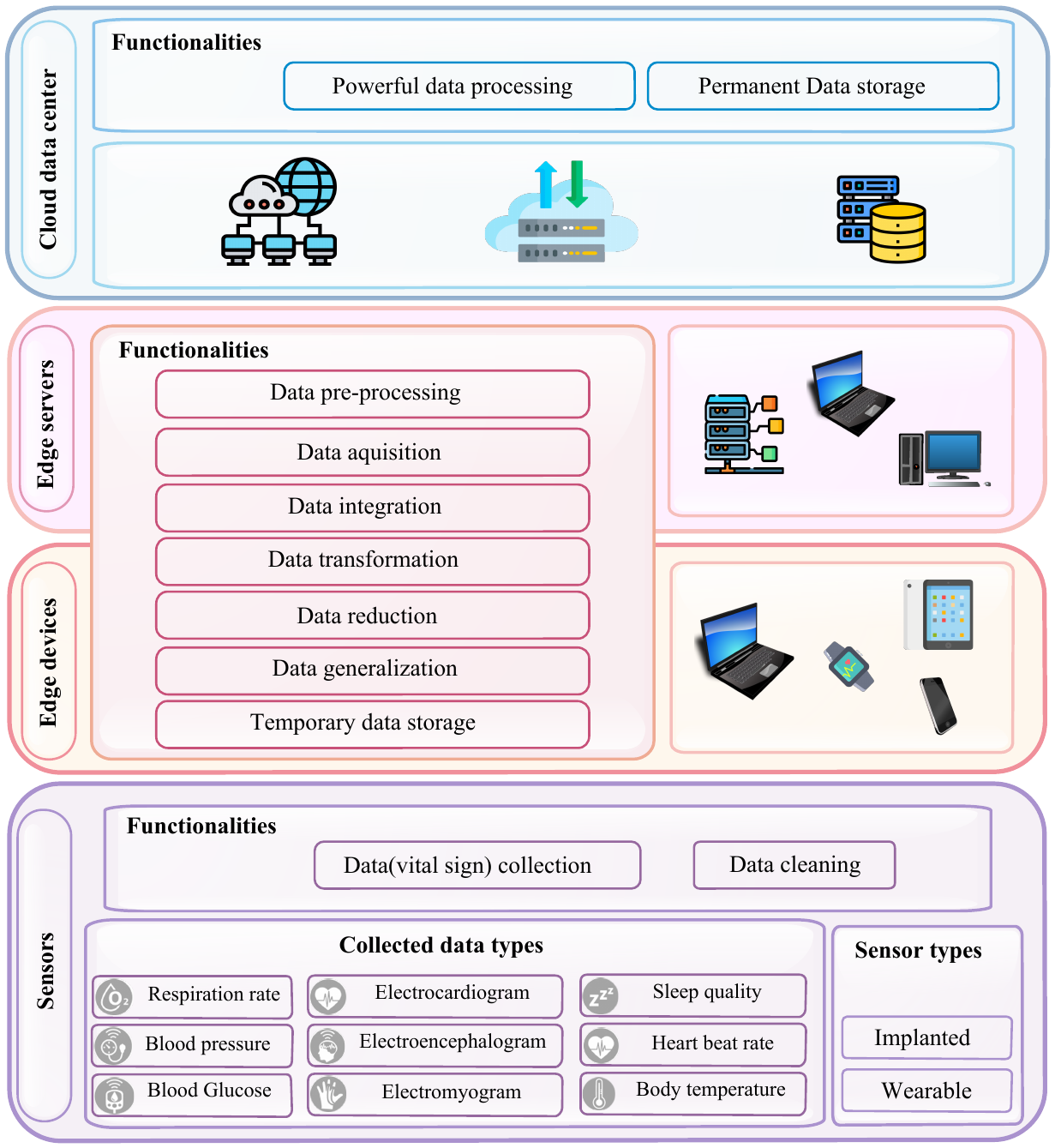}  
    \caption{Edge-assisted smart healthcare functionality flow.}
    \label{fig.arch2}
\end{figure}

    By and large, edge computing architecture in smart healthcare applications consists of four layers, including sensors layer, edge devices layer, edge servers’ layer, and cloud data center layer. Figure \ref{fig.arch2} illustrates this architecture as a functionality flow, which indicates the functionality and the common devices used in each layer. The sensor layer contains different sensors, including implanted and wearable sensors that are responsible for collecting vital signs and data from patients. We indicate some data types that can be collected via sensors in a smart healthcare system. The sensor layer communicates with the edge devices layer via short range and low power wireless communication protocols such as ZigBee, Bluetooth, and Wi-Fi \cite{azar2019energy}. Edge devices layer includes users’ devices such as smartphones, raspberry pi, tablets, and smart watches, which are capable of conducting some processes and short-term storing data. The information is sent via Bluetooth/ Wi-Fi/internet from edge devices layer to edge servers’ layer. Edge servers’ layer comprises micro data centers and is able to conduct more processes and stores more data in comparison to the edge devices layer \cite{abdellatif2019edge}. As shown in Figure \ref{fig.arch2}, the functionality of edge devices layer and edge servers layer are similar. The only difference is that the storage capacity and the computational power in edge server layer is more powerful than edge device layer.
    Cloud data centers’ layer encompasses more powerful processors and has the ability to store the vast amount of data. This layer communicates with edge servers’ layer via the Internet.
    
\section{AI in edge assisted smart healthcare systems}
\label{Sec.AI}
\begin{figure} [htbp!] 
    \includegraphics[height=85mm, width=90mm]{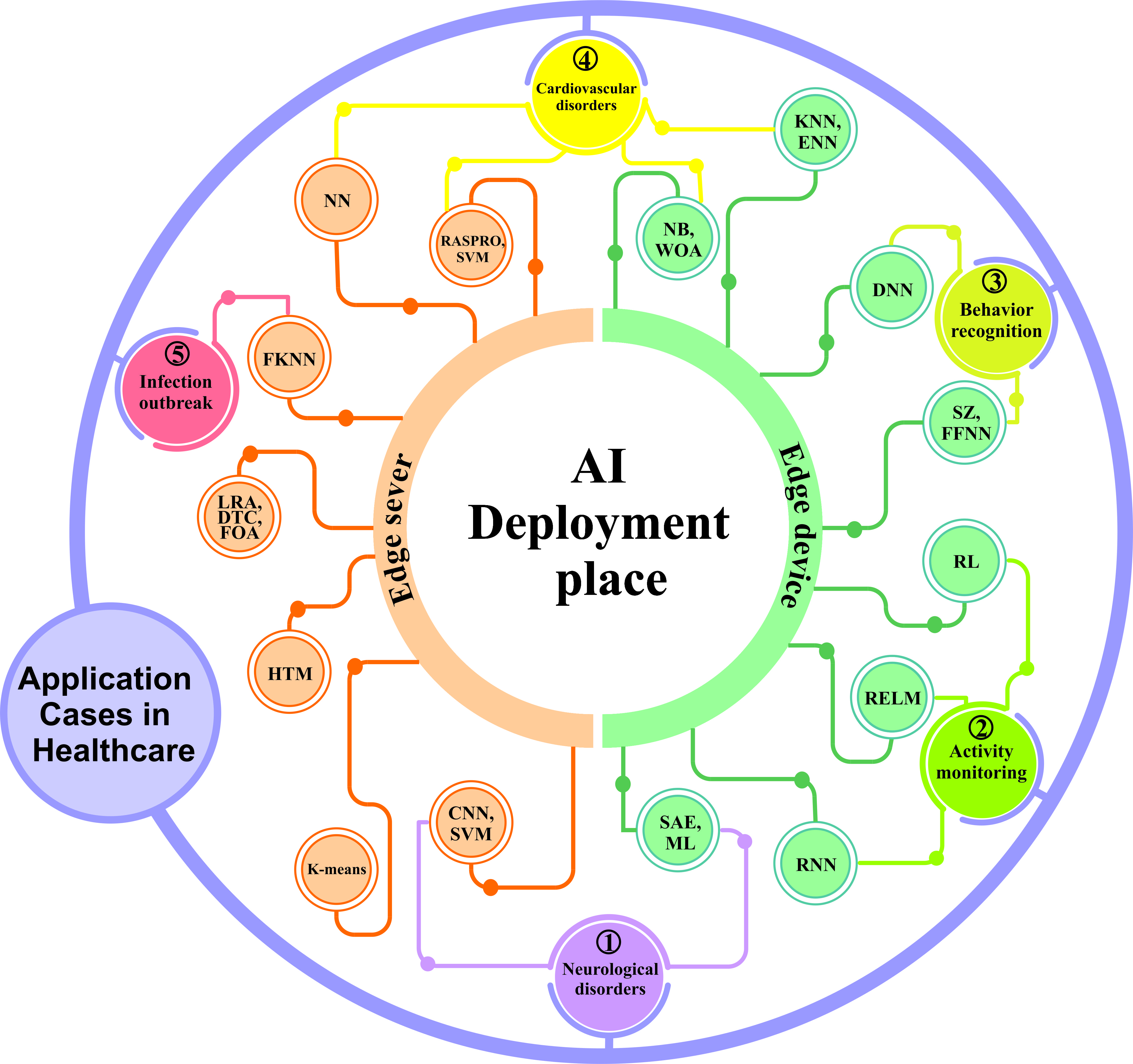}  
    \caption{The classification of state-of-the-art solutions based on the AI deployment place and their application case.}
    \label{fig.AI2}
\end{figure}

\begin{figure} [htbp!] 
    \centering
    \includegraphics[height=90 mm, width=90mm]{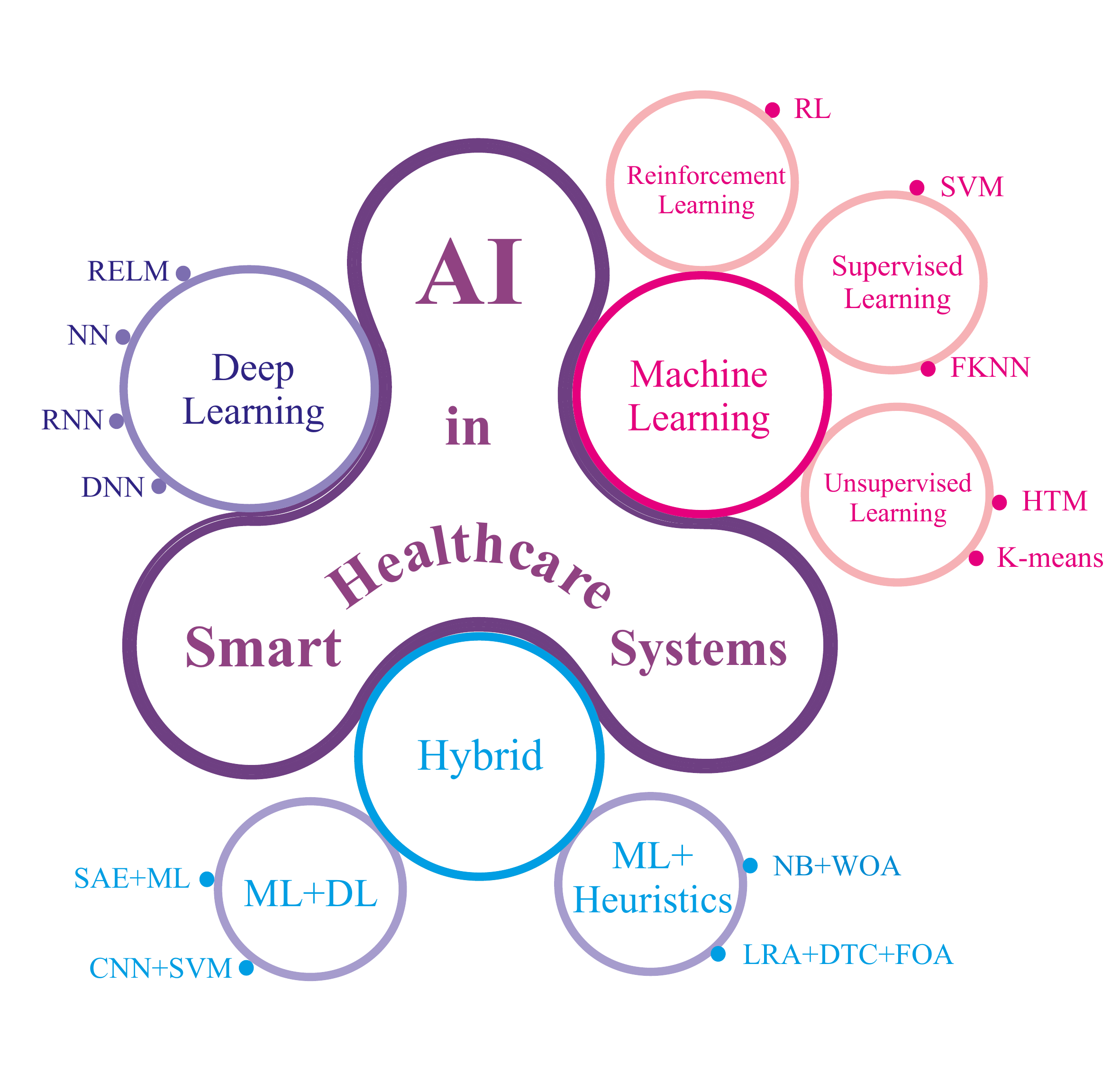}  
    \caption{The classification of state-of-the-art solutions based on the adopted AI techniques.}
    \label{fig.AI1}
\end{figure}

    This section first provides an overview and benefits of applying AI in edge assisted smart healthcare systems. Then, some related research works are examined in detail. Three different classifications of the investigated works are provided and illustrated in Figure \ref{fig.AI2} and Figure \ref{fig.AI1}. Finally, a brief discussion along with Table \ref{tab.sum} including the summary of inspected papers, are brought in this section.
    
    The first classification considers the AI deployment place in edge computing to classify the papers. Regarding this, papers are classified into two groups, including \textit{edge device deployment} and \textit{edge server deployment}. The second classification categorizes the research works based on their applied cases in smart healthcare into five groups (see column application case in Table \ref{tab.sum}). The mentioned five groups include \textit{1. Neurological disorders} (e.g., Epileptic seizure, chronic brain disorders), \textit{2. Activity monitoring }(e.g., Elderly monitoring, fitness tracking, emergency care, and tele-health advice), \textit{3. Behavior recognition} (e.g., stress detection, feeling recognition), \textit{4. Cardiovascular disorders} (e.g., Heart problems, blood pressure), \textit{5. Infection outbreak }(e.g., various viruses such as Chickungunya, COVID 19). These two classifications are outlined through Figure \ref{fig.AI2}, which illustrates the adopted algorithms, their associated healthcare cases, and their deployment places on edge. For instance, the DNN algorithm is employed for behavior recognition, which is deployed at an edge device. However, it can be seen that three of papers (CNN, SVM\cite{hosseini2020multimodal}, K-means \cite{shu2019mobile}, and HTM \cite{greco2019edge}) are not assigned to any of the application case groups, since their application healthcare cases have not been clarified by the authors. Moreover, Figure \ref{fig.AI1} depicts the third categorization, which is based on the adopted AI technique, including three subcategories: \textit{ML}, \textit{DL}, and \textit{hybrid} methods. The ML-based techniques, including \textit{supervised learning}, \textit{unsupervised learning}, and \textit{RL}, and also, hybrid technologies encompassing the hybrid of ML and DL, the hybrid of ML and heuristic methods.

\subsection{Overview and Benefits}
    The integration of AI and edge technology enables smart healthcare systems to run AI techniques locally on end-user devices where data are produced, while in traditional systems data processing procedures have been performed by AI techniques placed in the cloud far from end-users. This integration provides the possibility of processing data on edge devices in a few milliseconds, which delivers real-time results. Since healthcare services are directly related to human health, supplying real-time services and results is so crucial in critical conditions \cite{hassan2019hybrid}.
    
    Researchers have recently focused on moving the inference portion of the AI workflow to the edge devices. The smart healthcare system can conduct the training portion in one environment and run the other parts in another place. In order to train ML/DL, the vast amount of data and powerful computing items are needed; thus, the cloud is more suitable for this part of AI techniques. Whereas, the inference part or running the trained models in novel records can be executed at edge devices. Moreover, some techniques such as model compression methods by reducing large AI models into slight ones can move some training to the edge devices gradually \cite{azar2019energy}.
    
    All in all, the adoption of AI techniques for processing medical data on the edge of the network can contribute to providing smarter healthcare services with prompt response, high privacy, high accuracy, improved reliability, enhanced bandwidth usage, low power consumption and low latency. Also, it can reduce the burden on the network by diminishing the amount of transmitted data to the cloud servers \cite{azar2019energy}.

\subsection{Review and Classification}
    In this subsection, AI-based edge assisted smart healthcare solutions are discussed in two groups according to the first classification mentioned earlier (See figure \ref{fig.AI2}).

\subsubsection{AI in edge devices} \label{sec.AIinDevice}
    Researchers in \cite{al2019lightweight} have outlined a novel architecture for human activity recognition that utilizes Dockers technology on edge devices accompanied by AI techniques. The proposed architecture consists of sensors network, central cloud server controlling the sensors, and edge computing as a bridge between sensors and cloud. The regularized extreme learning machine (RELM) is used in edge computing to classify the collected data from the sensors.
    
    Also, in \cite{abdellatif2019edge}, the authors have proposed a context-aware edge-assisted smart healthcare framework by exploiting AI techniques. They have adopted DL for data compression in mobile edge devices to diminish bandwidth and energy usage. They have employed stacked auto-encoders (SAE), which is a type of neural network and hierarchically extracts data. Also, in order to guarantee high reliability and a swift responding in case of detecting or predicting disorders in epileptic seizure detection, the authors have adopted a data classification method leveraging ML algorithms.
    
    Researchers in \cite{uddin2019wearable} have proposed a human activity prediction system adopting edge technology and AI techniques. In the proposed system, data is collected by sensors and transferred to the edge device (laptop), where features are extracted from the collected data. Then, with regards to the features, a Recurrent Neural Network (RNN) is trained on the edge device. The trained RNN is utilized for predicting human activities. \vspace{-.5cm}
    
    Authors in \cite{rachakonda2019stress} have proposed a smart healthcare system using edge technology and DL for real-time stress level detection by taking physical activity into account. The proposed system adopts wrist band sensors to collect the required data. Also, DNN is utilized on edge devices. In the employed DNN, the input layer receives the collected data from sensors. Some hidden layers are responsible for data and stress level analyzing, the output layer reports the detected stress level, and eventually, the results are stored in the cloud. In the proposed network, a supervised learning method is adopted for data training, the gradient descent algorithm is employed for optimizing the training algorithm, and logistic regression is used for classifying of stress levels.
    
    Also, researchers in \cite{min2018learning} have designed an offloading model for edge-aided smart healthcare systems adopting AI methods to preserve user privacy in emergency care and Telehealth advice application. The proposed model uses the RL-based offloading algorithm to protect the privacy of healthcare data during the processing procedure. Researchers in \cite{djelouat2020real} have designed a real-time elderly monitoring system on a heterogeneous multi-core edge device using AI techniques. The proposed system utilizes the shimmer device for data acquisition; then, the collected data is compressed by a compressive sensing (CS) method. Next, the compressed data is transferred to the edge gateway through Bluetooth. The ODROID XU4 is used as an edge gateway where the compressed data is reconstructed and classified to identify various abnormal events.
    
    An energy-aware data reduction method form edge-assisted smart healthcare systems adopting AI methods is proposes in \cite{azar2019energy}. This system using body area network (BAN), wireless sensor network (WSN), and wearable devices to collect the vital signs. The fast error bounded lossy compression pattern (SZ) has been utilized for feature extraction in wearable devices. Afterward, the compressed data are transferred to a local PC for further processing and analysis. Researchers have deployed a part of DL on edge devices and the training network at the cloud. Moreover, a Feed-Forward Neural Network (FFNN) is deployed in an edge device (PC) to classify the compressed data, which helps with detecting the stress level for the case study. Researchers in \cite{hassan2019hybrid} have proposed a real-time smart remote monitoring framework using edge technology and AI methods. The proposed framework consists of four layers, including the sensing layer for data collection, edge device layer for offline pre-processing, edge server layer, and cloud layer for further online processes. The offline classification methods are trained in the cloud for predicting the health state of the patient in disconnected emergency events. The proposed method adopts whale optimization algorithm (WOA) and naïve Bayes classifier (NB) for choosing the slight set of features with high accuracy level.

\subsubsection{AI in edge servers} \label{sec.AIinServer}
    In \cite{pathinarupothi2018iot}, the authors have proposed an edge-based smart healthcare system to detect severity and transmit alerts for cardiovascular diseases. Leveraging ML-based algorithms like rapid active summarization for effective prognosis (RASPRO) and support vector machine (SVM), the proposed system has conducted data compression, data fusion, and data classification in an android smartphone as an edge device. Moreover, the authors in \cite{rani2018smart} have introduced a new model for controlling the outbreak of the Chickungunya virus. The proposed smart healthcare system leverages edge technology as a bridge between cloud storage and end-users. BANs and smartphones are adopted to collect data and transfer them to the edge servers where ML-based fuzzy k-nearest neighbor (FKNN) algorithm classifies the data coming from edge devices. The results are sent to cloud data centers. Also, in \cite{tuli2020healthfog}, the authors have proposed a new smart healthcare framework that utilizes edge technology and DL technique for automatically analyzing and predicting cardiac diseases. This system delivers a smart healthcare service as a fog service. The proposed model includes body area networks (sensors), edge gateways (smartphone), and Fog-Bus module that consists of broker node (Laptop), worker node (Raspberry Pi), and cloud data center (Microsoft Azure). The set partitioning in hierarchical trees (SPIHT) and singular value decomposition (SVD) techniques are adopted for data filtering in the broker node, and an ensemble DL technique is used for data classification in the worker node.

    Researchers in \cite{hosseini2020multimodal} have designed an automatic epileptic seizure prediction and localization system exploiting edge technology and AI techniques. They have proposed a novel multimodal data processing method on mobile edge computing environment using convolutional deep learning (CNN) and wavelet transform for extracting high order features and a nonlinear SVM classifier along with generalized radial basis function (GRBF) kernel and long short-term memory (LSTM) DL for classifying the extracted electroencephalogram (EEG) and DL features.
    
    Also, in \cite{greco2019edge}, a real-time data analysis solution for medical data stream analysis has been performed, which leverages edge technology and AI methods for detecting data anomaly. The proposed architecture comprises four different layers. The sensor layer fetches and dispatches medical data using wearable devices. The pre-process layer adopts raspberry pi to convert the collected raw medical data streams into resource description framework (RDF) streams. In the cluster processing layer, a hierarchical temporal memory (HTM) algorithm is employed to cope with the anomaly detection issue. The last layer, persistent layer stores the analyzed data to provide the possibility of further evaluation.
    
\begin{table*}[htbp!]
    \centering
    \caption{Summary of AI-based edge assisted smart healthcare solutions}
    \begin{tabular}{|m{0.02\textwidth}|m{0.1\textwidth}|m{0.1\textwidth}|m{0.12\textwidth} |m{0.1\textwidth} |m{0.1\textwidth}|m{0.11\textwidth}|m{0.22\textwidth}|}
    \hline
         \rotatebox[origin=c]{90} {\textbf{Reference}} & \textbf{Algorithm}  & 
         \textbf{AI Category} & 
         \textbf{AI contribution} & 
         \textbf{AI deployment place} &
         \textbf{Healthcare application case} &
         \textbf{Main objective of the system} & 
         \textbf{AI benefits}\\
    \hline
         \cite{al2019lightweight} & 
         RELM & 
         DL & 
         Data classification & 
         Edge device (Raspberry Pi) &
         Activity monitoring &
         Human activity detection & 
         Low communication latency, swift decision making, and low cost\\
    \hline
         \cite{abdellatif2019edge} & 
         DL(SAE) +ML & 
         Hybrid & 
         Data compression (feature extraction), data classification & 
         Edge device (smartphone) &
         Neurological disorders &
         Epileptic seizure detection & 
         Low bandwidth usage, low energy usage, low response time, and high reliability\\
    \hline
        \cite{uddin2019wearable} & 
        DL (RNN) & 
        DL & 
        Predicting the activities & 
        Edge device (laptop) & 
        Activity monitoring &
        Human activity prediction & 
        Real-time supporting and high prediction rate (99.69\%) \\
    \hline
        \cite{rachakonda2019stress} & 
        DL (DNN) & 
        DL & 
        Stress analysis & 
        Edge device (Smartphone) &
        Behavior recognition &
        Stress Level Detection & 
        High detection accuracy (99.7\%), rapid detection, and real-time supporting\\
    \hline
        \cite{min2018learning} & 
        RL & 
        RL & 
        Data privacy & 
        Edge device &
        Activity monitoring &
        Data privacy & 
        High privacy, low energy consumption, and low computational latency\\
    \hline
        \cite{djelouat2020real} & 
        CS, ENN/KNN & 
        Hybrid & 
        Data compression, data classification & 
        Sensors, Edge gateway (ODROID XU4) & 
        Cardiovascular disorders &
        Resource allocation & 
        Low execution time, high accuracy, low latency, and real-time supporting \\
    \hline
        \cite{azar2019energy} & 
        SZ algorithm, FFNN & 
        Hybrid & 
        Data compression and classification & 
        Edge devices (PC, wearable device) &
        Behavior recognition &
        Data reduction & 
        High lifetime, low energy consumption, and high accuracy\\
    \hline
        \cite{hassan2019hybrid} & 
        NB, WOA & 
        Hybrid & 
        Feature selection and data classification & 
        Edge devices & 
        Cardiovascular disorders &
        Remote monitoring & 
        Real-time supporting, big data analysis, and high accuracy\\
    \hline
         \cite{pathinarupothi2018iot} & 
         RASPRO, SVM & 
         Supervised & 
         Data compression,  data fusion, and  data classification & 
         Edge gateway & 
         Cardiovascular disorders &
         Severity detection and alerts transmission for cardiovascular diseases & 
         Low bandwidth usage, high availability, and low cost\\
    \hline
        \cite{rani2018smart} & 
        FKNN & 
        Supervised & 
        Data classification & 
        Edge servers & 
        Infection outbreak &
        Control the Chickungunya virus infection & 
        Real-time supporting, low execution time, and low latency\\
    \hline
        \cite{tuli2020healthfog} & 
        NN & 
        DL & 
        Data classification & 
        Worker node in Fog (raspberry pi) & 
        Cardiovascular disorders &
        Automatic Heart Disease analysis & 
        High prediction accuracy and low execution time\\
    \hline
        \cite{hosseini2020multimodal} & 
        CNN, SVM & 
        Hybrid & 
        Feature extraction, data classification & 
        Edge server &
        Neurological disorders &
        Data analysis & 
        High accuracy, high precision, high sensitivity, and reduce the network congestion\\
    \hline
        \cite{greco2019edge} & 
        HTM & 
        Unsupervised & 
        Data anomaly detection & 
        Edge server & 
        Not mentioned &
        Data analysis & 
        High accuracy\\
    \hline
        \cite{shu2019mobile} & 
        K-means & 
        Unsupervised & 
        Data dissemination & 
        Edge server &
        Not mentioned &
        Data dissemination & 
        High reliability, high accuracy\\
    \hline
        \cite{lin2018task} & 
        LRA, DTC, FOA & 
        Hybrid & 
        Resource allocation, classification the requests & 
        Edge server (cloudlet) & 
        Not mentioned &
        Resource allocation & 
        Low energy consumption, low execution time, and low cost\\
    \hline
    \end{tabular}
        \label{tab.sum}
\end{table*}

    In \cite{shu2019mobile}, authors have designed an edge-based data dissemination model in which the mobile edge servers are utilized for disseminating various data segments. The proposed model includes the installation method and data propagation technique that are respectively responsible for ensuring the integrity of data objects and preventing the transmission collision. Moreover, in order to provide efficient data dissemination method in heterogeneous edge networks, researchers have used adaptive protocol and the k-means clustering algorithm in this research.

    Researchers in \cite{lin2018task} have suggested a task offloading and resource allocation model for edge-assisted smart healthcare systems using AI methods. The authors have considered edge computing with three layers, including devices layer, edges layer, and clouds layer. Cloudlets with micro data centers are deployed at the edge of the network as edge servers. The proposed model uses a linear regression algorithm (LRA) at the edge servers to analyze the user requests and a pipeline tree classifier (PTC) for clustering the jobs. Moreover, fruit fly optimization algorithm (FOA) is used to optimize the resource allocation procedure.
\subsection{Discussion and Summary}
    With regard to the investigated papers, most of the researchers have adopted AI on edge technology to manage the collected medical data. In most cases, data management contributes to reducing the amount of transmitted data. Since a lower amount of data needs to be conveyed to the cloud, cost, response time, latency, and network congestion will be decreased, bandwidth and energy usage will be diminished, and accuracy and reliability will be enhanced. Also, the smart healthcare system will be able to make decisions swiftly. The last column of Table \ref{tab.sum} provides more information about AI benefits in edge assisted smart healthcare solutions. Th ensuing section discusses a new system model which tries to provide a single system covering most of the mentioned AI contributions and benefits.

\section{System model} \label{sec.frame}

\begin{figure*} [htbp!]
   \centering
    \includegraphics[height=129 mm, width=160 mm]{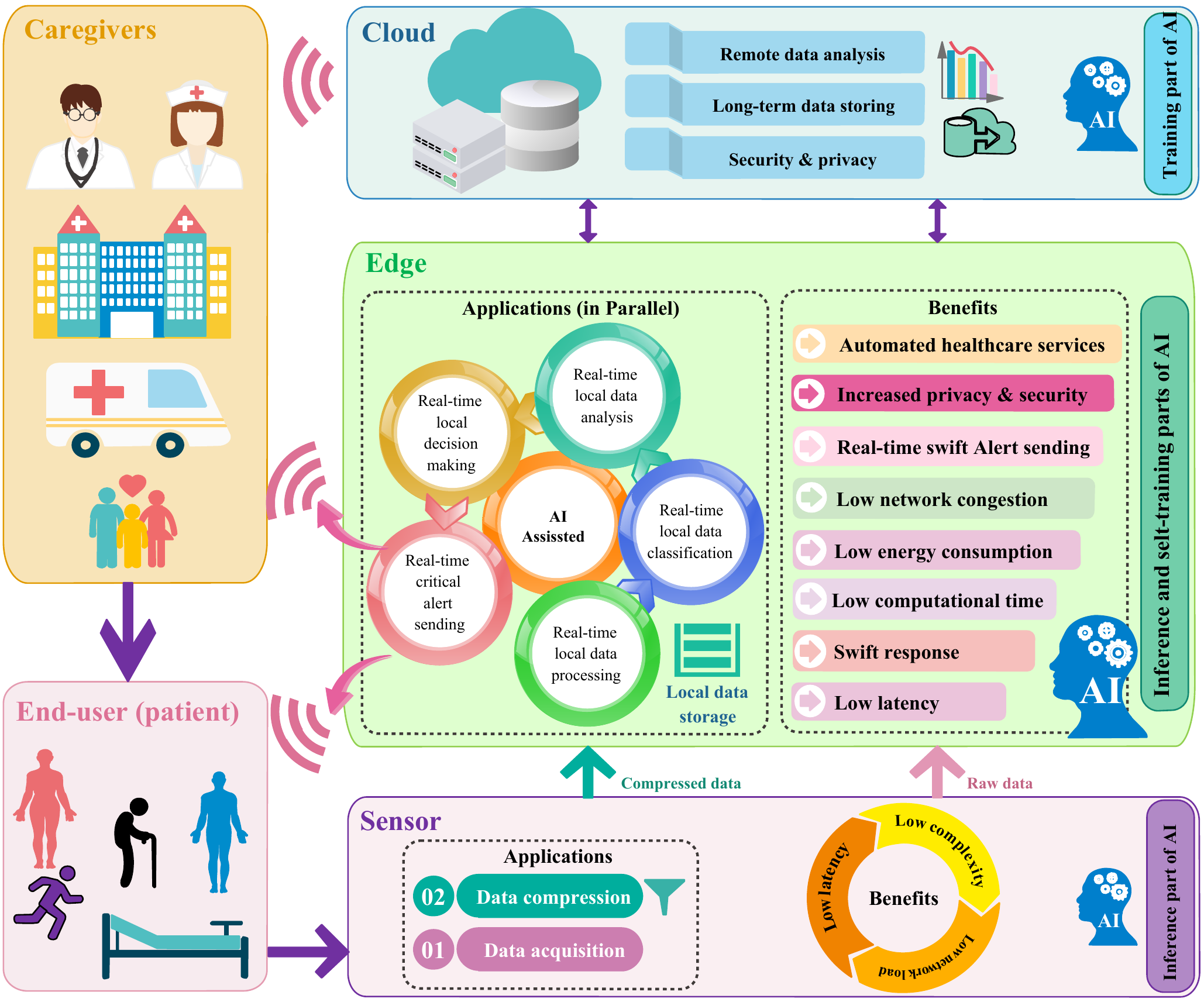}
    \caption{An Overview of the System Model}
    \label{fig.sm}
\end{figure*}

    This section describes a novel smart healthcare framework leveraging edge technology and employing different AI techniques like DRL. Figure \ref{fig.sm} demonstrates the proposed system model with three main layers, including \textit{sensor layer}, \textit{edge layer}, and \textit{cloud layer}. In this model, the end-user can be a patient at a smart home being monitored remotely, a patient in a health institute, elderly people, or even a healthy person who cares for his fitness. Caregivers can be any of a health professional, a health institute, a mobile ambulance, family members, or a drugstore.
    
    The proposed system adopts different AI techniques, including DRL, distributed from sensors to the cloud, to facilitate smart healthcare services. Being flexible, the proposed framework can automatically define a proper solution map for the encountered issue. Based on the size of the faced problem and the volume of possible collected data, it can distribute the processes from sensors to cloud or from edge to cloud. Regarding this, various layers of DRL can be deployed on different layers of the proposed framework. As a case in point, should the purpose of the system is fitness tracking for a healthy person, there is no need to perform part of processes on the cloud. Therefore, the system can accomplish processes on sensors and edge layer. The results can either be temporarily stored on edge or permanently on the cloud according to the end-user’s request. The system is flexible enough to choose the appropriate place for hosting the input and output layers of DRL, depending on the current healthcare issue.
    
    The sensor layer is typically responsible for collecting data from the end-user and transferring the raw records to the edge layer, either to the edge devices or directly to the edge servers. We believe that compressing data in the sensor layer helps the system to reduce the overall latency, network congestion, and computation complexity on the edge layer. The data compression can be handled by adopting lightweight AI techniques such as CS on the sensors.
    
    The edge layer consists of edge devices and edge servers hosting AI techniques. Many research attempts have been done in the field of smart healthcare leveraging AI and edge technology. However, there are still some unaddressed challenges. For example, DL needs a vast volume of data to train the algorithms. But, in some cases such as Tele-surgery, there is not enough data available to be trained. Also, the trained data for a specific disease using DL is not adoptable for another similar illness; thus, the data should be re-trained. Moreover, RL, as an autonomous self-training system, should make several small decisions to achieve a sophisticated objective. In case of complex decisions with many states, like monitoring a mobile patient with a chronic disease which may have countless states due to the unstable context information of a moveable patient, the RL algorithm may not be able to handle the problem since learning from all states and regulate the reward path can be uncontrollable for the algorithm. A potential solution for these limitations can be an AI technique that uses both estimation and mapping the states together. In fact, the proposed framework can use a real-time self-training model or training from datasets depending on the encountered circumstances. This characteristic makes the system flexible to be used for various issues based on the demands of end-users, like remote monitoring, self-monitoring, or clinical monitoring. Table \ref{tab.AItech} summarizes the pros and cons of the common AI techniques utilized in the smart healthcare domain. Hence, considering the result of Table \ref{tab.AItech} and the discussion, DRL is seen as the best choice since it can cover the cases mentioned above. According to the issue size, the system can automatically decide to map all the states/ use estimation or both together.
    
    In case of critical health issues, real-time alert sending and delivering health services in-time is so crucial to save a person’s life. Hence, reducing latency, and time of decision making are vital in smart healthcare systems. The above mentioned flexible AI technique, DRL, can be adopted in parallel on the edge layer by distributing data and processes on various devices or different hidden layers of DRL. It is crystal clear that the distributed parallel model contributes to more reduction of latency, and more swift decision making in comparison to an unparalleled model. Since the processes are performed faster than the prior methods, the power consumption can also be diminished noticeably. Therefore, the lifetime of battery-powered devices will be increased.
    
    The medical data, either compressed or raw, are transferred from sensors to edge layer where they are locally stored, classified, and analyzed adopting an AI technique in parallel mode to deliver real-time decisions for critical healthcare requests. Soon after making a decision, the corresponding alert would instantly be sent to the caregivers/end-user or both. Thus, they can take action in time and prevent a possible catastrophe. Exploiting autonomous and smart self-training AI methods lied on edge technology, the proposed model can offer an automated smart healthcare system that can properly work even without human intervention.
    
      \begin{table*}[]
   \caption{Merits and demerits of common AI techniques in smart healthcare domain}
    \centering
  
    \begin{tabular}{|l|m{0.3\textwidth}|m{0.5\textwidth}|}
    
    \hline
         \textbf{AI Technique} & 
         \textbf{Benefits} & 
         \textbf{Drawbacks} \\
    \hline
         \textbf{ML} & 
         Improves over time, Adapt without human intervention & 
         Needs high amount of time and resources for learning, High level of error susceptibility\\
    \hline
        \textbf{RL} & 
        High learning speed, Real-time response, Dynamic, Capability of learning from small data &
        Lack of clear interpretability, Lack of Integration of Prior Knowledge\\
    \hline
        \textbf{DL} &
        High dimensional (with many layers), Handle large amounts of data &
        It just provides  approximate statistics not accurate results, Needs tons of data for training, Low speed evolution, Data quality (medical data are highly ambiguous and heterogeneous; thus, training phase may face some problems like data sparsity, redundancy, and missing value), Transferring it to the other problems is not easy\\
    \hline
        \textbf{DRL} & Attain complex smart healthcare states, best-accuracy, powerful model, scalability, complex decision-making, & Potential dead-zones, may incur complexity states, off-policy learning, heavy computations, continuous data adjustment \\
    \hline
    \end{tabular}
    \label{tab.AItech}
 \end{table*}
    
    Furthermore, considering the critical nature of IoT devices and medical data which are vulnerable to be attacked by hackers, security and privacy management are in the list of vital concerns in smart healthcare systems and should be taken into account. In our vision, security and privacy management are considered in different levels distributed from sensor to cloud layer adopting automated AI techniques like DRL. The proposed system model can cover crucial issues like network security, data security, data privacy, user privacy, communication privacy, encryption, and authentication of users to prevent illegal access to the system.
    
    Eventually, the cloud layer performs remote and further data analysis and long-term data storing. With regards to the constraint resources of IoT and edge devices, in case of too complex processes, the cloud can be the best place for conducting further analysis that is not possible to be done on edge. For instance, should the smart healthcare system to train a vast amount of data, the cloud can be the best choice for this purpose.
    
    In summary, the smart components of the proposed system model adopt sensors to collect data related to the real-time patients’ status while all integrated with physical items. All the components are connected to a cloud-oriented system which collects and processes all the data produced by sensors. The proposed system model can be considered as a digital twin system which adopts real world and real-time data to perform simulation leveraging edge intelligence and wireless communication, that can estimate how a process can be established to get the better results in terms of predicting a disease in advance or saving lives in critical circumstances. Moreover, the proposed system model has a reasonable complexity since there are not any complex algorithms used in the model. The overall complexity of the proposed system model depends on the health issue and the amount of data. However, it cannot be greater than the complexity of DRL itself. 

\section{Challenges and Future Research Directions} \label{sec.openissues}
    According to our investigation, the ensuing open issues and challenges in the adoption of AI in edge-assisted smart healthcare solutions are detected. Future research directions in this area include both technical and non-technical challenges summarized as follows. The first three challenges are considered as non-technical and the rest of challenges are known as the technical challenges. 
    
  \subsection{Sensor Usage} The most important issue in terms of sensor adoption is how patients comfort in using sensors. Sensors can be either implanted or wearable. The elderly people may have difficulties in using wearable ones. Some others may refuse to adopt the implanted sensors. This challenge also implies a personal comfort zone, in other words, what seems comfortable to one patient cannot be assumed the same for another. 
    
    \subsection{Alert Receiving} The most salient concern in alerting is how family members/trusted relatives/caregiver are reliable and responsive in taking reaction once they get an alert. For instance, they may not get the alerting notification in time due to different reasons (e.g., the dead battery of their device, being busy with other tasks, being far from the patient). These issues may result in irreparable losses. Thus, fault tolerance measures need to be considered. 
    
     \subsection{Patient's concerns} Some patients are concerned about the treatment procedure and how they can trust a digital system for distinguishing or predicting their diseases. Thus, they usually resist using smart technologies, since they do not have enough information about how the system works and how doctors interact with the system and indirectly with the patients.
    
   \subsection{Data Loss} Although efficient data compression contributes to reducing the amount of transmitted data, which may diminish the latency, power usage, and bandwidth usage, data integrity and information loss as a result of data compression is still a crucial open issue for smart healthcare solutions.
   
   \subsection{Lightweight AI Techniques} Since edge computing apparatuses, which are mostly used in smart healthcare applications, consist of constraint resources and limited power supply, there is a need for lightweight AI techniques that can be executed on them. Although AI techniques usually contribute to obtaining high accuracy, in some cases, it is achieved at the expense of augmented computational cost. An effective resource utilization may lead to the enhancement of lightweight AI techniques, which can help the smart healthcare system by analyzing enormous volumes of medical records produced continuously. RL techniques, in comparison to other AI methods, need lower memory resources and computational complexity. The RL techniques usually learn from their prior experiences; then, they do not need training with datasets. Thus, using RL can lead to preserving computational resources and time as well, which are considered as the critical criteria in the healthcare systems.
     
   \subsection{Redundant Training Data} The DL techniques adopt a huge sample space of datasets for training. Each sample consists of various features that are adopted in the detection/ prediction of diseases. Nevertheless, there is substantial redundancy in samples transferring almost identical information. Hence, the computational resources should work on processing similar information many times, which increases the time and computational complexity. These two factors are so essential in case of real-time critical medical situations to save the patient's life. Therefore, proposing a strong and efficient algorithm for diminishing redundant training and providing effective resource utilization still is a challenging issue.
     
    \subsection{Privacy and Security} Since data is stored close to the end-users thanks to the edge technology, individual users own the data entirely and can protect them. However, due to the constraint resources of users’ devices, they cannot benefit from strong privacy and security protection algorithms. Thus, there is still considerable room for enhancing the privacy and security of smart healthcare systems leveraging edge technology and AI. A lightweight AI technique for coping with this issue can be a promising research area.
     
    \subsection{Automatic Network Management} With regard to the critical nature of smart healthcare systems, it is so vital for these systems to have the capability of real-time and automatically managing the network issues. The adoption of lightweight AI techniques such as lightweight RL can lead to smart and robust network traffic management in smart healthcare systems. Also, in order to prevent the security gaps and learn the network for autonomously diminishing any possible attacks, the smart healthcare system can utilize an ML-based intrusion detection system. Moreover, software-defined network (SDN) has a crucial role in the implementation of the autonomous network management. The use of learning algorithms can help to control the virtualization of network components and functions to deliver a self-regulated smart healthcare system.
    
\section{Conclusion}\label{sec.conclusion}
    This article illuminates how edge technology, along with AI techniques, can enhance the smart healthcare systems. Processing locally, edge technology improves smart healthcare systems by reducing latency, network burden, and power consumption. Also, AI injects the smartness to the system. The integration of AI and edge technology makes the components of smart healthcare systems smarter and brings many advantages to them. We introduce a novel smart healthcare framework that leverages different AI techniques in a parallel mode on edge technology. The proposed model distributes the processes from sensors to the cloud servers. Conducting some processes using a lightweight AI technique on sensors results in diminishing the latency, complexity, and network load. Furthermore, adopting parallel processing on edge layer employing a flexible AI method contributes to more reduction of latency, swift response and decision making, and sending real-time alerts to the caregivers, which are so vital in smart healthcare services. The proposed model considers security and privacy requirements for the entire smart healthcare system, including the sensitive medical data, the associated network connections, and users’ access to the smart healthcare system. Although the proposed system model covers some limitations of prior smart healthcare frameworks, there are still other unaddressed challenges that need to be addressed, comprising data loss, and autonomous network management.

\balance
\bibliographystyle{IEEEtran}
\bibliography{bib}


\end{document}